\documentclass[12pt,journal,compsoc,onecolumn]{IEEEtran}
\usepackage{amsmath,epsfig}
 \usepackage{lingmacros}
 \usepackage{tree-dvips}
\usepackage{amsmath} 
\usepackage{amssymb}
\usepackage{pifont}
\usepackage[noadjust]{cite}
\usepackage{multirow}
\newcommand{\cmark}{\ding{51}}
\newcommand{\xmark}{\ding{55}}

\newlength{\bibitemsep}
\newlength{\bibparskip}
\let\oldthebibliography\thebibliography
\renewcommand\thebibliography[1]{
  \oldthebibliography{#1}
  \setlength{\parskip}{\bibitemsep}
  \setlength{\itemsep}{\bibparskip}
}
\def\x{{\mathbf x}}

\def\XX{{\bf X}}

\def\Y{{\cal Y}}
\def\S{{\cal I}}

\def\x{{\bf x}}
\def\y{{\bf y}}
\def\D{{\cal D}}

\def\F{{ f}}
\def\DD{{d}}
\def\tr{{\bf tr}}
\def\1{{\bf 1}}
\def\I{{\cal I}} 
\def\p{{p}}
\def\q{{q}}
\def\x{{\bf x}}  
\def\Y{{\cal  Y}}  
\def\V{{\bf V}}
\def\Z{{\bf Z}}
\def\G{{\cal G}}

\usepackage[ruled,vlined]{algorithm2e}

\title{Label-frugal satellite image change detection with generative virtual exemplar learning}
\author{Hichem Sahbi \\
$ $ \\
Sorbonne University, CNRS, LIP6,  F-75005, Paris, France 
 }

\begin{document}
 \maketitle
\begin{abstract}
  Change detection is a major task in remote sensing which consists in finding all the occurrences of changes in multi-temporal  satellite or aerial images. The success of existing methods, and particularly deep learning ones, is tributary to the availability  of hand-labeled training data that capture the acquisition conditions and the subjectivity of the user (oracle). In this paper, we devise a novel change detection algorithm, based on active learning. The main contribution of our work resides in a new model that measures how important is each unlabeled sample, and provides an oracle with only the most critical samples (also referred to as virtual exemplars) for further labeling. These exemplars are generated, using an invertible graph convnet, as the optimum of an adversarial loss that (i) measures representativity, diversity and ambiguity of the data, and thereby (ii) challenges (the most) the current change detection criteria, leading to a better re-estimate of these criteria in the subsequent iterations of active learning. Extensive experiments show the positive impact of our label-efficient learning model against comparative methods.
\end{abstract}
\section{Introduction}
\label{sec:intro}

\noindent Satellite image change detection is nowadays becoming a hotspot in remote sensing with applications ranging from anthropogenic activity monitoring to phenology mapping, and natural hazard damage assessment~\cite{ref4,ref5,ref5v1,refffabc4}. However, this task is known to be challenging due to the pervasive changes in satellite images resulting from sensors and acquisition conditions (weather variations, radiometric changes, scene intrinsic content, etc). Early work relies on comparisons of multi-temporal series by analyzing how changes unfold over time using image differences and thresholding, spectral vegetation indices, principal component and change vector analysis~\cite{ref7,ref9,ref11,ref13,ref13v1,ref13v2}.  Other methods either (i) rely on normalization as a preprocessing step that minimizes the effects of variations due to atmospheric conditions (e.g., clouds, haze,...) and sensor calibration~\cite{ref14,ref15,ref17,ref20,refffabc8}, or (ii) model the expected appearance of relevant changes (objects or regions of interest over time) using statistical machine learning~\cite{ref21v1,ref21,ref25,ref26,refffabc7,refrefrefICIP2014,ref28,ref27,refffabc3,refffabc0} while accounting for irrelevance changes.\\

\indent The success of machine learning models, particularly deep neural networks \cite{refffabc1,sahbi2021b}, depends on the availability of large collections of hand-labeled reference images that capture the diversity of relevant and irrelevant changes \cite{refffabc5,refffabc6}. These collections should also capture the user's targeted changes. However, hand-labeling large data collections is time-consuming and impractical, and even when relatively tractable, may suffer from domain-shift and may also struggle to capture the nuances of the user's subjectivity and intention. Several works explore solutions to make machine learning more efficient and less reliant on large collections of labeled data including few-shot and transfer learning \cite{reff45v1,reff45}, as well as semi/self-supervised learning \cite{refff2v1,refff2}. Nonetheless, these methods often lack the ability to directly incorporate and understand the user's intention.  Solutions based on active learning  \cite{reff1,reff2,reff13,reff16,reff15,reff53,reff12,reff74,reff58,refsahbicvpr2008} are rather more appropriate and consist in efficiently probing the user about the relevance of {\it only the most informative} observed changes prior to train decision criteria that best suit the user's intention and scene acquisition conditions.\\

\indent In this paper, we propose a new label-efficient satellite image change detection algorithm based on invertible graph convnets. The proposed model is interactive and queries the oracle about the labels of the {\it most critical} data (also dubbed as virtual exemplars) whose positive impact on the trained convnets is the most important. These virtual exemplars are designed as the most representative, diverse and ambiguous data that optimize an adversarial loss. This loss is optimized while learning an invertible graph convnet that achieves both classification and generation so as {\it to constrain the synthesized virtual exemplars to lie on a nonlinear manifold enclosing both ``change'' and ``no-change'' classes}. Note that the solution presented in this work, despite being adversarial, is conceptually different from the ones widely used in generative adversarial networks (GANs) \cite{refffabc}. Indeed, whereas GANs aim at generating {\it fake} data that mislead the trained discriminators, our formulation seeks instead to generate {\it critical} data --- for annotation --- which are the most impactful on the subsequent learned classifiers. Put differently, the proposed method allows to sparingly query the oracle {\it only} on the most representative, diverse and uncertain data which challenge the current discriminator, and ultimately lead to more accurate ones in the subsequent iterations of change detection.  Extensive change detection experiments  show the effectiveness of our invertible graph convnets and virtual exemplar learning models against comparative methods.

\section{Proposed method}
\label{sec:format}

Given two  satellite images  $\I_r = \{\p_1, \dots , \p_n\}$, $\I_t = \{\q_1, \dots , \q_n\}$  as a collection of registered patch-pairs taken at two different instants  $t_0$, $t_1$, with $p_i$ and $q_i \in \mathbb{R}^d$. Our goal is to train a convnet $f$ that predicts the unknown labels $\{\y_i\}_i$ in $\I=\{\x_i=(p_i,q_i)\}_i$ with $\y_i=1$ iff $\x_i$ corresponds to a change pair, and  $\y_i=0$ otherwise. As training $f$ requires patch-pairs (hand-labeled by an oracle), we seek to make $f$ as {\it label-efficient} as possible and also  accurate.

\subsection{A glimpse on interactive change detection}
Our change detection algorithm relies on a question \& answer (Q\&A) process that iteratively probes the oracle about the labels of the most {\it critical patch-pairs}, and then updates change detection criteria accordingly. The most critical patch-pairs, at iteration $t$, constitute a {\it display} $\D_t$ whose unknown labels are denoted as $\Y_t$. As shown subsequently, our algorithm trains change detection criteria $f_0,\dots,f_{T-1}$ iteratively, starting from a random display $\D_0$, and according to the following steps \\

\noindent i) Gather $\Y_t$ from the oracle and train $f_t (.)$ on $\cup_{k=0}^t (\D_k, \Y_k)$ using graph convolution networks (GCNs)~(see section~\ref{inv}).\\
\noindent ii) Select the subsequent display  $\D_{t+1}\subset \S\backslash\cup_{k=0}^t \D_k$ to show to the oracle. 
A strategy considering all configurations of displays $\D \subset \S\backslash\cup_{k=0}^t \D_k$, prior to training the underlying criterion $f_{t+1} (.)$ on $\D \cup_{k=0}^t \D_k$, and keeping only the display  $\D$ with the highest positive impact on $f_{t+1} (.)$'s accuracy, is {\it clearly intractable}. Besides, this requires labeling each of these configurations by the oracle. Our contribution introduced in this paper (see~\ref{formulation},~\ref{inv2}) is based instead on active learning strategies which are rather more appropriate. However, one should be cautious as many of these strategies are found to be equivalent to (or worse than) basic random data selection strategies (as also discussed in  \cite{reff2} and references therein).\\

Considering the aforementioned goal, the main contribution of this paper includes a novel display selection strategy that finds in a {\it flexible way} virtual exemplars instead of using rigid unlabeled data (in contrast to \cite{refff33333}). The design principle of our method allows selecting the most  diverse, representative, and uncertain data that challenge (the most) the current change detection criteria and leads to a better re-estimate of these criteria in the subsequent iterations of active learning. Besides, our contribution also includes a novel invertible GCN design that achieves both classification and virtual exemplar generation more effectively as shown through extensive experiments on the challenging task of interactive satellite image change detection.  
\subsection{Virtual exemplar design}\label{formulation}
As labeling images in $\I$ is highly prohibitive, one should focus only on the most critical samples. The latter define the subsequent display ${\cal D}_{t+1}$, also referred to as virtual exemplars $\{\V_k\}_{k=1}^K$, used to train $f_{t+1}(.)$ on ${\cal D}_{t+1} \cup \dots \cup {\cal D}_{0}$. We consider for each training sample $\x_i \in \I$ a conditional probability distribution  $\{\mu_{ik}\}_k$ that measures the membership of $\x_i$ to the $K$-virtual exemplars in $\{\V_k\}_{k=1}^K$. These memberships $\mu=\{\mu_{ik}\}_{ik}$ together with the underlying virtual  exemplars are found by minimizing the following objective function 
{
\begin{equation}\label{eq01}
\hspace{-0.25cm}\begin{array}{ll}
\displaystyle   \min_{\V; \mu \in \Omega}    & \displaystyle   \tr\big (\mu \ \DD(\V,\XX)^\top \big)  \  + \ \alpha  \ [\1^\top_n  \mu] \log [\1^\top_n \mu]^\top \\
                 &  \  +  \  \beta \ \tr\big(\F(\V)^\top \  \log \F(\V)\big) \ + \ \gamma \ \tr(\mu^\top \log \mu),                    
\end{array}
\end{equation}}

\noindent being  $\Omega=\{\mu :  \mu  \geq 0; \mu \1_K = \1_n\}$ a convex set that constrains $\mu$ to be row-stochastic, and   $\1_{K}$, $\1_{n}$ are two vectors of $K$ and $n$ ones respectively. In Eq.~\ref{eq01}, $\tr(.)$ is the matrix trace operator, $\mu \in \mathbb{R}^{n \times K}$ is a learned matrix whose i-th row  corresponds to the conditional probability of assigning $\x_i$ to each of the $K$ learned virtual exemplars in $\V \in \mathbb{R}^{d \times K}$, and $\log$ is applied entry-wise. In the above equation, the matrix $\DD(\V,\XX) \in \mathbb{R}^{K \times n}$ captures the euclidean distances between the virtual exemplars in $\V$ and the input data in $\XX$ whilst $\F(\V) \in \mathbb{R}^{\#Classes \times K} $ corresponds to the softmax layer. The first term in Eq.~\ref{eq01} measures the representativity of the virtual exemplars by constraining them to be {\it close} to their assigned input data in $\I$ and thereby inheriting more accurate labels when the oracle annotates their closest input data. The second term in Eq.~\ref{eq01} models diversity by maximizing the spread of the probability distribution of samples across the learned virtual exemplars; this term reaches its minimum when virtual exemplars attract {\it diversely} input data in $\I$. The third term measures the {\it ambiguity} (or uncertainty) in $\V$ as the negative of the softmax entropy, and it reaches its smallest value when virtual exemplars in  $\V$ are evenly scored w.r.t different classes. Finally, the fourth term acts as a regularizer which considers that without any a priori about the three other terms, the membership distribution $\{\mu_{ik}\}_k$ should be uniform for each sample $\x_i$. All these terms are combined using $\alpha, \beta, \gamma \geq 0$.\\

\indent In order to solve the optimization problem in Eq.~\ref{eq01} for each change detection cycle  $t$, we consider a bi-level optimization procedure that fixes the display $\V$ and finds $\mu$, and vice versa. One may show that Eq.~(\ref{eq01}) admits the following solution 
\begin{equation}\label{eq2}
\begin{array}{lll}
  \mu^{(\tau+1)}& :=&\displaystyle  \textrm{\bf diag} \big(\hat{\mu}^{(\tau+1)} \1_K\big)^{-1} \  \hat{\mu}^{(\tau+1)} \\
 \V^{(\tau+1)} &:= & \hat{\V}^{(\tau+1)} \  \textrm{\bf diag} \big(\1^\top_n {\mu}^{(\tau)} \big)^{-1},
 \end{array}  
\end{equation}
being  $\hat{\mu}^{(\tau+1)}$, $\hat{\V}^{(\tau+1)}$ respectively
\begin{equation}\label{eq3}
\begin{array}{l}
  \exp\big(-\frac{1}{\gamma}[\DD(\XX,\V^{(\tau)}) + \alpha (\1_n \1^\top_K + \1_n \log \1^\top_n \mu^{(\tau)} )]\big),\\
  \\
  \XX \ \mu^{(\tau)}+ \beta \sum_{c} \nabla_v f_c(\V^{(\tau)}) \circ  (\1_d \ [\log f_c(\V^{(\tau)})]^\top + \1_d \1_K^\top),
\end{array} 
 \end{equation}
 here $\textrm{\bf diag}(.)$ maps a vector  to a diagonal matrix and $\circ$ is the Hadamard matrix product. Due to space limitation, details of the proof are omitted and  result from the optimality conditions of Eq.~\ref{eq01}'s gradient. Initially, ${\mu}^{(0)}$ and  ${\V}^{(0)}$ are set to random values and the procedure converges to an optimal solution in few iterations, defining the most relevant virtual exemplars of $\D_{t+1}$ used to train the subsequent GCN $f_{t+1}$ (see algorithm~\ref{alg1}).
 
\begin{algorithm}[!ht]
\footnotesize 
  \KwIn{$\S$, ${\cal D}_0 \subset {\S}$, budget $T$.}
\KwOut{$\cup_{t=0}^{T-1} (\D_t,\Y_t)$ and $\{f_t\}_{t}$.}
\BlankLine
\For{$t:=0$ {\bf to} $T-1$}{$\Y_t \leftarrow \textrm{oracle}(\D_t)$; \\ 
  $f_{t} \leftarrow \arg\min_{f} {\textrm{CrossEntropyLoss}}(f,\cup_{k=0}^t (\D_k,\Y_k))$; \\
 $\tau \leftarrow 0$; $\hat{\mu}^{(0)} \leftarrow \textrm{random}$;  $\hat{\V}^{(0)} \leftarrow \textrm{random}$;\\
Set ${\mu}^{(0)}$ and  ${\V}^{(0)}$ using Eqs.~(\ref{eq2}) and (\ref{eq3});
\BlankLine
 \While{($\footnotesize \|\mu^{(\tau+1)}-\mu^{(\tau)}\|_1 +  \|\V^{(\tau+1)}-\V^{(\tau)}\|_1 \geq\epsilon  \newline \wedge  \tau<\textrm{maxiter})$}{
   Set ${\mu}^{(\tau+1)}$ and  ${\V}^{(\tau+1)}$ using Eqs.~(\ref{eq2}) and (\ref{eq3}); \\  
   $\tau \leftarrow \tau +1$;
 }
$\tilde{\mu} \leftarrow \mu^{(\tau)}$;  $\tilde{\V} \leftarrow \V^{(\tau)}$; \\ 
 ${\D_{t+1} \leftarrow \big\{\x_i \in \S\backslash \cup_{k=0}^t \D_k: \x_i\leftarrow\arg\min_\x\|\x-\V_k\|_2^2 \big\}_{k=1}^K}$.
}
\caption{Virtual exemplar learning}\label{alg1}
\end{algorithm}
 \subsection{Graph convnets}\label{inv} 
\def\VV{{\cal V}}
\def\E{{\cal E}}
\def\A{{\bf A}}
\def\UU{{\bf U}} 
\def\W{{\bf W}}
\def\N{{\cal N}}
\def\SS{{\cal S}}
\def\FF{{\cal F}}

Given a graph $\G=(\VV, \E)$ as a set of nodes $\VV$ and edges $\E$ associated to a given patch-pair $\x \in \I$. The graph $\G$ is endowed with a signal\footnote{In change detection, this signal is the difference between patch-pair pixel values.} 
and associated with a learnable adjacency matrix $\A$. Graph convnets (GCNs) seek to learn a set of $C$ filters  $\FF$ that define convolution on $n$ nodes of $\G$ (with $n=|\VV|$) as $(\G \star \FF)_\VV = g_2\big(g_1(\A \  \UU^\top) \W\big)$, here $\UU \in \mathbb{R}^{s\times n}$  is the  graph signal, $\W \in \mathbb{R}^{s \times C}$  is the matrix of convolutional parameters corresponding to the $C$ filters and  $g_1$, $g_2$ are nonlinear activations applied entry-wise. In  $(\G \star \FF)_\VV$, the input signal $\UU$ is first mapped using $\A$ and this defines for each node $u$, the set of neighbor  aggregates. Beside learning the convolution parameters in $\W$, entries of $\A$ are also learned in order to capture an ``optimal'' graph topology when achieving aggregation so as $(\G \star \FF)_\VV$ models both attention and convolution layers; the first layer aggregates signals in $\N(\VV)$ (sets of node neighbors of $\VV$) by multiplying $\UU$ with $\A$ while the second layer achieves convolution by multiplying the resulting aggregates with the $C$ filters in $\W$. Stacking one or multiple attention+convolution layers $(\G \star \FF)_\VV$ together with fully connected and softmax layers defines the whole architecture of our GCNs.   In the remainder of this paper, we go one step further, and make these GCNs bijective, i.e., invertible. As shown subsequently, this property is valuable, as it allows us {\it to learn the virtual exemplars in a mapping (latent) space while capturing the topology of their manifold in the input (ambient) space.} This turns out to be more effective when learning display models as shown later in experiments.  

\subsection{Invertible graph convnet design}\label{inv2}
Subsequently, we subsume the aforementioned GCN architecture as a multi-layered neutral network $f_{t,\theta}$ with $t$ being the active learning cycle, $\theta=\{\W_1,\dots,\W_L\}$, $L$ its depth, $\W_\ell \in \mathbb{R}^{d_{\ell-1} \times d_\ell}$ its $\ell^{th}$-layer weight tensor, and $d_\ell$ the output dimension of the $\ell^{th}$-layer. In what follows, we omit both $\theta$ and $t$ in the definition of $f_{t,\theta}$ and we rewrite the output of a given layer $\ell$ of  this GCN as $\phi^\ell=g_\ell (\W_\ell^\top \phi^{\ell-1})$, $\ell \in \{2,\dots,L\}$, being $g_\ell$ an activation function; without a loss of generality, we also omit the bias in the definition of $\phi^\ell$. 
\noindent Provided that the matrices in $\theta$ are invertible, and the activation functions $\{g_\ell\}_\ell$ bijective\footnote{In practice, $\{g_\ell\}_\ell$ are chosen as leaky-ReLU activations which are bijective.}, one may guarantee that the trained GCN is also bijective and henceby invertible. Besides, the dimensionality of all the layers remains constant\footnote{Excepting the softmax layer whose dimensionality depends on the number of classes. Nonetheless a simple trick consists in adding fictitious softmax outputs to match any targeted dimensionality.}. This bijection  property is valuable, as it reduces the complexity of solving Eq.~\ref{eq01} by finding virtual exemplars in the latent space instead of the ambient space. On another hand, by the invertibility of the GCN, this guarantees that the inverted latent exemplars belong to the nonlinear manifold/distribution enclosing data in the ambient space. \\ 

\noindent By considering the matrix of virtual exemplars, denoted as $\Z$ in the latent space, and by the invertibility of $\F$ (i.e., $\F(\V) = \F(\F^{-1}(\Z))=\Z$),  Eq.~\ref{eq01} can be rewritten as
\begin{equation}\label{eq02}
\hspace{-0.25cm}\begin{array}{ll}
\displaystyle   \min_{\Z; \mu \in \Omega}    & \displaystyle   \tr\big (\mu \ \DD(f^{-1}(\Z),\XX)^\top \big)  \  + \ \alpha  \ [\1^\top_n  \mu] \log [\1^\top_n \mu]^\top \\
                 &  \  +  \  \beta \ \tr\big(\Z^\top \log \Z\big) \ + \ \gamma \ \tr(\mu^\top \log \mu).                    
\end{array}
\end{equation}
This objective function can be solved w.r.t. $\Z$ (similarly to Eqs.~\ref{eq01} and~\ref{eq2}) using the fixed-point iteration process, while recovering $\V=\F^{-1}(\Z)$ thanks to the invertible GCN. Hence, this surrogate problem allows exploring nonlinear manifolds in the ambient space while being tractable in the latent space.

\def\III{{\bf I}}
\subsection{Regularization}
The success of the generative properties of the invertible GCN $f^{-1}$ is reliant on the stability of $f$. In other words, when $f$ is $M$-Lipschitzian (with $M \approx 1$), the network  $f^{-1}$ will also be $M$-Lipschitzian (with $M\approx 1$) \cite{refffabc888},  so any slight update of virtual exemplars in the latent space (with the fixed-point iteration) will also result into a slight update in these exemplars in the ambient space when applying $f^{-1}$. This eventually leads to stable virtual exemplar generation in the ambient space, i.e., they follow the actual distribution of data manifold. As the Lipschitz constant of $f$ is $M=\prod_{\ell} \|\W_\ell \|_2 . \big|g'_\ell\big|$, the sufficient conditions that guarantee that both $f$ and $f^{-1}$ are $M$-Lipschitzian (with $M\approx 1$) corresponds to (i) $\|\W_\ell \|_2 \approx 1$, and (ii) $\big|g'_\ell\big| \approx 1$ for all $\ell$. Hence, by design, conditions (i)+(ii) could be satisfied by choosing the slope of the activation functions to be close to one (in practice to 0.99 and 0.95 respectively for the positive and negative parts of the leaky ReLU), and also by constraining the norm of all the weight matrices to be {\it orthonormal} which also guarantees their invertibility. This is obtained by adding a regularization term, to the cross-entropy (CE) loss, when training GCNs, as
\begin{equation} 
\min_{\{\W_\ell\}_\ell}{\textrm{CE}}(f;\{\W_\ell\}_\ell) + \lambda \ \sum_{\ell}\big\|\W_\ell^\top \W_\ell-\III\big\|_F,
\end{equation}
here $\III$ stands for identity, $\|.\|_F$ denotes the Frobenius norm and $\lambda>0$ (with $\lambda=\frac{1}{d}$ in practice); in particular, when $\W_\ell^\top \W_\ell-\III =0$,  then $\W_\ell^{-1}=\W_\ell^\top$ and  $\|\W_\ell \|_2 =\|\W_\ell^{-1} \|_2=1$.  With the aforementioned formulation, the learned GCNs are guaranteed to be discriminative, stable and invertible.
\section{Experiments}
We evaluate change detection  performances using the  Jefferson dataset which consists of  $2,200$ non-overlapping and registered patch-pairs with each patch including $30\times 30$ pixels in RGB.  These registered pairs correspond to a large area of Jefferson (Alabama) captured by two (bi-temporal) GeoEye-1 satellite images of $2, 400 \times  1, 652$ pixels taken at two different instants (in 2010 and in 2011) with a spatial resolution of 1.65m/pixel. These images capture a few relevant changes (road network damages, building destruction, etc., due to tornadoes that hit this area in 2011) together with no-changes (irrelevant ones such occlusions due to clouds, and radiometric variations). In total, the Jefferson dataset includes 39 positive pairs (relevant changes) and  2,161 negative pairs, so less than 2\% of the observe area correspond to relevant changes, and this makes relevant changes rare and difficult to find. In all experiments, half of the dataset is used to train our display and GCN models whilst the other half is used for evaluation.  As the two (positive and negative) classes are highly imbalanced, we score the performance of change detection, on the eval set, using the equal error rate (EER). In the observed results, smaller EERs imply better performances.  
\subsection{Ablation Study}
In this section, we study the impact of different terms of our objective function individually, pairwise  and all jointly taken. For all the settings, the last term of Eq.~\ref{eq01} is kept as it allows obtaining the closed form in Eq.~\ref{eq2}.
Table~\ref{tab1} shows the impact of each of these terms and their combination; from these results, we observe an important impact of representativity and diversity particularly at the earliest change detection iterations. The impact of ambiguity is more noticeable in the later iterations as it further refines change detection criteria when all the modes of ``change'' and ``no-change'' distribution are explored. The EERs are shown for increasing sampling percentages defined, at each iteration $t$, as $(\sum_{k}^{t} |\D_k|/(|\I|/2))\times 100$  with again $|\I|=2,200$ and $|\D_k|$ set to $16$. 
 \begin{table} \caption{This table shows an ablation study of our display model. Here rep, amb and div stand for representativity, ambiguity and diversity respectively. These results are shown for different iterations $t$ (Iter) and the underlying sampling rates (Samp) again defined as $(\sum_{k}^{t} |\D_k|/(|\I|/2))\times 100$. The AUC (Area Under Curve) corresponds to the average of EERs across iterations.}\label{tab1} 

 \resizebox{1.01\columnwidth}{!}{
 \begin{tabular}{ccc||ccccccccc||c}
 rep  & div & amb &2 & 3& 4& 5& 6& 7& 8& 9 & 10 & AUC. \\
 \hline
 \hline
        \xmark  &  \xmark &     \cmark       & 27.29  &  11.15  &  7.97 &   8.18  &  7.31  &  7.97  &  7.94 &   7.50  &  7.90 &  10.35  \\
         \xmark  & \cmark  &       \xmark     & 18.72  &  11.24&     7.97 &   8.18&     7.29&     7.59   &  7.88   &  7.50  &   7.90 & 9.36   \\
         \cmark   &  \xmark  &      \xmark    &  35.98 &   16.86 &   6.52 &    4.98&     2.67 &    2.03   &  1.80 &    1.45&     1.30 & 8.17 \\
\hline
         \cmark   &      \xmark &   \cmark     &  40.40&   23.86&     9.56&    7.65&     5.75 &    5.47  &   6.12 &    4.40   &  5.72 & 12.10 \\
         \xmark  &  \cmark &   \cmark     & 27.29  &   11.15    & 7.97  &   8.18   &  7.31   &  7.97   &  7.94  &   7.50  &   7.90 &   10.35 \\
         \cmark  & \cmark    &     \xmark    &  29.84 &    17.63 &    6.21  &   4.40   & 2.70  &   1.98 &    1.92  &   1.65   &  1.52 & 7.53 \\
\hline 
   \cmark   &  \cmark  &  \cmark   &  27.61  &   11.76 &    5.74  &   2.95  &   2.39  &   1.89  &   1.61  &   1.55 &    1.34 &\bf 6.31  \\
\hline \hline 
\multicolumn{3}{c||}{Samp\%}  & 2.90 & 4.36& 5.81& 7.27& 8.72& 10.18& 11.63& 13.09 & 14.54 & - 
\end{tabular}}
\end{table}
\subsection{Comparison}\label{compare}

\begin{figure}[tbp]
  \begin{center} 
    \includegraphics[width=0.70\linewidth]{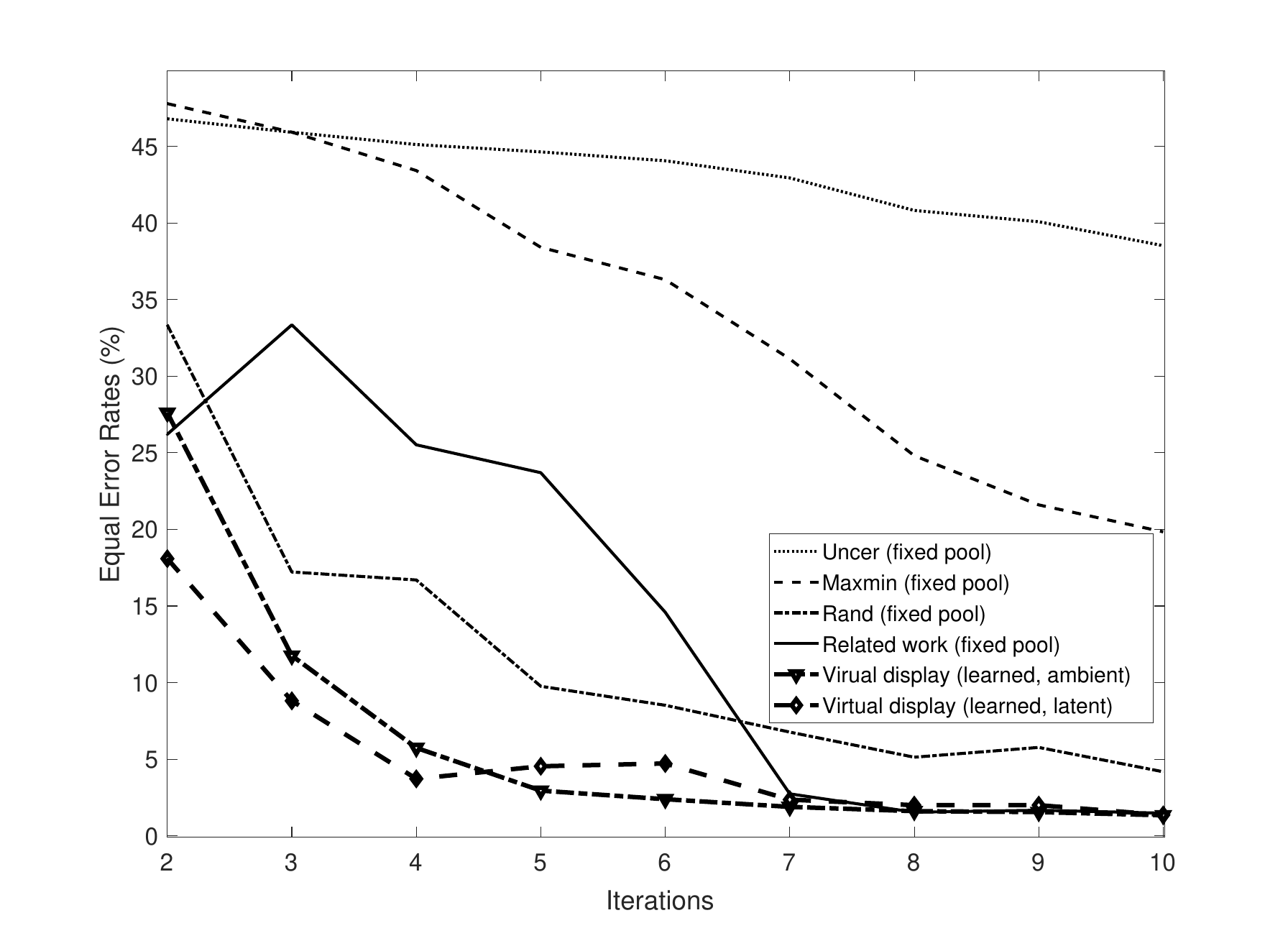}
    \end{center}
 \caption{This figure shows a comparison of different sampling strategies w.r.t. different iterations (Iter) and the underlying sampling rates in table~\ref{tab1} (Samp). Here Uncer and Rand stand for uncertainty and random sampling respectively. Note that fully-supervised learning achieves an EER of $0.94 \%$.  Related work stands for the method in \cite{refff33333}; see again section~\ref{compare} for more details.}\label{tab2}\end{figure}

 We further compare the strength of our virtual exemplar (display) model against other display selection strategies such as  {\it random, maxmin and uncertainty}. Random sampling consists in taking a random subset from the pool of unlabeled training data while uncertainty aims at picking the display whose classifier scores are the most ambiguous (i.e., similar through classes) on the same pool. Maxmin consists in greedily selecting data in $\D_{t+1}$ where each data $\x_i \in \D_{t+1} \subset \S\backslash \cup_{k=0}^t \D_k$ is taken  by {\it maximizing its minimum distance w.r.t.  $\cup_{k=0}^t \D_k$}, resulting into  $\D_{t+1}$ with the most distinct samples. We also compare our sampling method  against the closely related method \cite{refff33333} which consists in defining relevance measures on the whole unlabeled set and picking the display with the highest relevance. As an upper bound, we also show performances using the fully supervised setting which trains a single classifier on the full training set whose labels are taken from the ground-truth. Figure~\ref{tab2}  shows the positive impact of the proposed display model against the aforementioned sampling strategies across increasing iterations (and sampling rates). From these observations, most of the comparative methods (excepting \cite{refff33333}) are unable to find the change class accurately; at the early iterations, both random and maxmin capture the diversity without being capable of sufficiently refining change detection results at the subsequent iterations, whereas uncertainly allows us to refine change detection results but without enough diversity. The method in \cite{refff33333} has the advantage of gathering the advantage of all these comparative strategies, but suffers at some extent from the rigidity of the selected display (which is taken from a fixed pool). In contrast to all these methods, our virtual exemplar  model allows us to learn flexible displays particularly when these displays are constrained in the latent space with our invertible GCNs, particularly at frugal labeling regimes. 

\section{Conclusion}
In this work, we introduce a new interactive change detection technique built on top of active learning.  Our flexible display models constrained by our invertible convnets allow training highly label-efficient change detection criteria. The adversarial design of our display model enables effective selection of the most informative (representative, diverse and ambiguous) data that challenge (the most) the current change detection criteria and further improve the subsequent ones. Extensive experiments conducted on the challenging task of satellite image change detection show that the proposed virtual display model outperforms other sampling strategies.


\begin{thebibliography}{1}
\bibitem{aaaaa9}  H. Sahbi.  A particular Gaussian mixture model for clustering and its application to image retrieval.   Soft Computing 12 (7), 667-676
\end{thebibliography}
\end{document}